\documentclass[conference]{IEEEtran}
\usepackage{cite}
\usepackage{amsmath,amssymb,amsfonts}
\usepackage{algorithmic}
\usepackage{graphicx}
\usepackage{textcomp}
\usepackage{xcolor}

\usepackage{url}
\usepackage{xspace}
\usepackage{multirow}

\newcommand{\embed}{\mathcal{E}}
\newcommand{\corpus}{\mathcal{T}}
\newcommand{\algo}{\textsl{Embedding Fusion}\xspace}

\newcommand{\citep}{\cite}

\usepackage{amsthm}
\newtheorem*{remark}{Remark}

\def\BibTeX{{\rm B\kern-.05em{\sc i\kern-.025em b}\kern-.08em
    T\kern-.1667em\lower.7ex\hbox{E}\kern-.125emX}}
\begin{document}

\title{Fusing Vector Space Models for Domain-Specific Applications}

\author{\IEEEauthorblockN{Laura Rettig}
\IEEEauthorblockA{\textit{eXascale Infolab} \\
\textit{University of Fribourg}\\
Fribourg, Switzerland \\
laura.rettig@unifr.ch}
\and
\IEEEauthorblockN{Julien Audiffren}
\IEEEauthorblockA{\textit{CMLA, CNRS, ENS Paris-Saclay} \\
\textit{Paris-Saclay University}\\
Cachan, France \\
audiffren@cmla.ens-cachan.fr}
\and
\IEEEauthorblockN{Philippe Cudr\'e-Mauroux}
\IEEEauthorblockA{\textit{eXascale Infolab} \\
\textit{University of Fribourg}\\
Fribourg, Switzerland \\
philippe.cudre-mauroux@unifr.ch}
}

\maketitle

\begin{abstract}
We address the problem of tuning word embeddings for specific use cases and domains. 
% In this paper, we propose a method for combining several domain-specific word embeddings in order to improve the performance of an application. 
% We show that using general-purpose word embeddings, such as those built on top of the entire English wikipedia corpus, pose limitations when applied to specific tasks. 
We propose a new method that automatically combines multiple domain-specific embeddings, selected from a wide range of pre-trained domain-specific embeddings, to improve their combined expressive power.
%We therefor train word embeddings on domain-specific text corpora.
%From these specific embeddings, multiple are useful to the task at hand, and as such, we combine multiple embeddings by concatenating their word vectors for the same terms and running dimensionality reduction to preserve the efficiency of the original embedding dimensions.
Our approach relies on two key components: 1) a ranking function, based on a new embedding similarity measure, that selects the most relevant embeddings to use given a domain and 2) a dimensionality reduction method that combines the selected embeddings to produce a more compact and efficient encoding that preserves the expressiveness. 
%In order to determine the best word embeddings for a task, we propose selecting suitable embeddings via a ranking function according to the task at hand. 
%This way, given any application, we are able to find the most suitable combination of embeddings such that we achieve a better performance of the task compared to general-purpose word embeddings.
%We evaluate our approach using a state-of-the-art word embedding use case -- sentiment analysis of review texts -- to demonstrate the suitability of our ranking approach and the improvement delivered by the combined domain-specific embeddings.
We empirically show that our method produces effective domain-specific embeddings that consistently improve the performance of state-of-the-art machine learning algorithms on multiple tasks, compared to generic embeddings trained on large text corpora.
\end{abstract}


\section{Introduction}
\label{sec:introduction}
Word embedding techniques such as word2vec \cite{Mikolov:2013} have become a key building block of many NLP applications. These techniques capture the semantic similarities between linguistic terms based on their distributional properties from large textual contents and allow to easily represent words or phrases in a low-dimensional vector space that can be leveraged by downstream applications (e.g., language translation or sentiment analysis). 

As training word embedding models is computationally intensive and requires large text corpora, pre-trained models such as those provided by Facebook Research\footnote{fastText pre-trained word vectors: \url{https://fasttext.cc/docs/en/crawl-vectors.html}} or Google\footnote{word2vec models from Google's Code Archive: \url{https://code.google.com/archive/p/word2vec/}} are widely used today. These models are trained on very large unlabeled text corpora of billions of words and provide high-quality embeddings for a variety of languages. 

Despite the convenience they bring, using such readily-available, pre-trained models is often suboptimal in vertical applications \cite{hamilton2016inducing,Xu2018LifelongDW}; as these models are pre-trained on large, non-specific sources (e.g., Wikipedia and the Common Crawl for FastText, news articles for Google's word2vec models), they often cannot capture important semantic information from specific sub-domains or applications. Indeed, many words carry specific meaning depending on their context, which can be difficult to capture from generic or encyclopedic contents only. In addition, particular domains may use specific vocabulary terms that are not present in pre-trained embeddings. 

%\item While we could train the embedding on the task corpus, we still like the idea of pre trained embeddings : bring more information into the task (like transfer learning). 
% pcm: I find the transfer learning argument interesting, would be good to add it

Retraining word embeddings for a specific context is possible, though it is also extremely costly, both in terms of computational power and in the availability of large quantities of text that have to be fed into the model. Instead, we suggest in this paper a third solution combining the convenience of pre-trained embeddings with the effectiveness of dedicated models. The main intuition behind our method is to leverage a collection of domain-specific embeddings and to efficiently \emph{combine} them in order to capture the peculiarities of a given application domain as closely as possible. While sensible, this approach also raises two new challenges that have to be tackled: i) How to automatically select the most adequate embeddings from a library of pre-trained embeddings for a specific task and ii) how to efficiently and effectively combine different embeddings in order to obtain high-quality embeddings suitable for the task at hand.

We take on these challenges through a new method introduced in this paper that we call \algo: We solve the first challenge by introducing a ranking technique based on a comparative analysis of the language used in the task and in the embeddings text corpora. We tackle the second problem by using dimensionality reduction methods that combine the selected embeddings to produce a more compact and efficient -- yet similarly expressive -- encoding.

Our main contributions are hence as follows:
\begin{enumerate}
    \item  To the best of our knowledge, we are the first to propose the idea of \textit{dynamically selecting} and \emph{combining} several word embeddings to better capture a particular application domain;
    \item We introduce a new technique to rank word embeddings, based on their relevance to a particular domain;  
    \item We describe several techniques to combine a set of domain-specific embeddings into a new encoding better suited for the task at hand;
    \item Finally, we demonstrate through a comprehensive empirical evaluation on multiple test corpora that the use of our approach -- i.e., the combination of automatically selected and relevant domain-specific embeddings -- leads to consistently improved results compared to the use of generic embeddings.
\end{enumerate}

\section{Method}
\label{sec:method}
As discussed above, the key idea of \algo is to replace a general-purpose embedding by a carefully constructed combination of domain-specific embeddings. 
Our method relies on two main ingredients, namely
i) a principled algorithm to automatically select the most relevant embedding(s), depending on the task at hand, and 
ii) an efficient dimensionality reduction algorithm that combines the previously selected embeddings into a single embedding of fixed dimension $d$. These two contributions are presented in Sections~\ref{subsec: ranking} and \ref{subsec: combination}, respectively.

%Our main method consists of two parts: (1) Combining the word embeddings that are most suitable to the task at hand in a way that is both efficient and accurate, for which we need to (2) rank the suitability of the available word embeddings based on their training corpora according to the test corpus.

\subsection{Ranking Embeddings}\label{subsec: ranking}
In the following, we use the terms \emph{embedding} or \emph{encoding} interchangeably to refer to word embeddings. Encodings are characterized by their two core features: the learning algorithm used for training (see e.g. \cite{mikolov2013efficient,bojanowski2016enriching}), and the text corpus used as training set (such as Wikipedia).
While the choice of the learning algorithm has a significant impact on the resulting embedding, in this work we are more interested in the influence of the associated text corpus and more precisely the characteristics of the corpus (topic, type of language, etc.).
 Therefore, throughout this paper, we use the exact same algorithm for learning \emph{all} text encodings, namely word2vec~\cite{mikolov2013efficient} with the CBOW model and a window size of 5, outputting embeddings of dimension $d=300$, which is a common choice in state-of-the-art word embeddings. % TODO: justify choice of CBOW
  The text corpora are then the only difference between the embeddings and we use them to define the notions of similarity and ranking of pairs of encodings.

\paragraph{Domain-Specific Embeddings} We say that an encoding is domain-specific if its associated text corpus is almost 
exclusively constituted of documents pertaining to a specific topic. Examples of domain-specific embeddings that we use in our subsequent evaluations include \textit{drugs}, an embedding constructed on Wikipedia articles relating to pharmaceuticals, or \textit{twitter}, an embedding constructed from a series of tweets. %(domain-specific in its language). 
In theory, the use of domain-specific embeddings suited to the task at hand entails significant advantages: the presence of specialized words in the vocabulary such as colloquial expressions and a better encoding of homonyms, e.g., when given words may have a distinct meaning in a specific domain, such as \textit{calculus} that refers to kidney stones in a medical context rather than to a branch of mathematics.

% \paragraph{}

However, one significant drawback of domain-specific embeddings is that by definition, each encoding may only be suitable for a narrow range of tasks.
To solve this issue we provide a generic solution that can automatically analyze multiple domain-specific contents and evaluate their relevance to a given task. 
More precisely, we propose a new approach that uses both i) a large library of pre-trained domain-specific embeddings and
ii) a new similarity function to automatically score the relevance of each encoding of the library with respect to the corpus of the target task, resulting in a ranking of their usefulness.
The similarity function is introduced below.

\paragraph{Notations} Let $\big(\embed_i\big)_{i=1}^N$ denote a collection of $N$ different domain-specific embeddings. 
For each encoding $\embed_i$, we denote as $\corpus_i$ its corresponding text corpus containing $\vert \corpus_i \vert$ words. For any word $w \in \corpus_i$, let $\embed_i(w)$ designate the embedding of $w$ in $\embed_i$ and $\# \corpus_i(w)$  the number of occurences of $w$ in $\corpus_i$. 
In the following and without loss of generality, all the embeddings are assumed to have the same dimensionality $d=300$. 
Finally, let $\corpus$ be the text corpus of the target task and $\embed$ be a general-purpose embedding method.

\paragraph{Text corpus as distribution} In order to define a similarity between text corpora, we proceed as follows. 
First, we associate with each word $w$ in each corpus $\corpus_i$ a weight $\alpha_i(w)$ that reflects its relative frequency:
\begin{equation}\label{eq: def weight word}
\alpha_i(w)= \frac{\# \corpus_i(w)}{ \sum_{j=1}^N \# \corpus_j(w)  }   \frac{ \sum_{j=1}^N \vert \corpus_j \vert  } {\vert \corpus_i \vert }
\end{equation}
$\alpha_i(w)$ can be seen as the ratio between the frequency of the word $w$ in $\corpus_i$ and its frequency among the collection of all text corpora. Therefore, $\alpha_i(w)$ increases with how \emph{specific} $w$ is to $\corpus_i$; a word $w$ that is frequent in $\corpus_i$ but rare in the other text corpora will have a large weight $\alpha_i(w)$.
The idea is that such a word might be representative of the domain specific to $\corpus_i$ and therefore is associated with a higher weight.

Then, each text corpus is seen as a distribution over the space of embeddings of words, where the probability $p$ of each word $w$ is directly proportional to its weight.
Formally, given a \emph{generic} encoding $\embed$ (i.e., a general-purpose embedding used to represent all words $w$), $\corpus_i$ induces a discrete distribution $P_i$ over $\mathbb{R}^d$, such that
\begin{equation}\label{eq: def weight proba}
\begin{aligned}
\forall w \in& \corpus_i,
&P_i(\embed(w)) = \frac{\# \alpha_i(w)}{\sum_{w' \in \corpus_i} \# \alpha_i(w')}. 
\end{aligned} 
\end{equation}
The reasoning behind this definition is twofold. First, words with large weights should give insight into the corpus topic~\eqref{eq: def weight word}, while words with small weights may be the result of errors (e.g., a document that is wrongly associated with the corpus). Therefore, we set the probabilities to reflect this relative importance.
Second, we use embeddings instead of raw words, as embeddings naturally include a notion of similarity between words \citep{bojanowski2016enriching}. 
The generic embedding is used here to provide a common encoding for all words of each text corpus, as opposed to embeddings that are specific to a topic.

\paragraph{Similarity between text corpora} 
We define the similarity between two text corpora as the similarity between their respective probability distributions over $\mathbb{R}^d,$ using the RBF kernel. In other words, 
\begin{equation}\label{eq: def similarity corpora}
\begin{aligned}
s(&\corpus_i,\corpus_j) \doteq s(P_i,P_j) \\ 
&= \sum_{w \in \corpus_i} \sum_{w' \in \corpus_j} P_i(\embed(w))   P_j(\embed(w')) \times\\
 &\quad\quad\quad\quad \exp ( - \frac{  \| \embed(w) - \embed(w') \|_2^2}{\sigma^2} )
\end{aligned} 
\end{equation}
where $\sigma$ is the bandwidth of the RBF kernel (in this work we used $\sigma=0.01$), and $P_i(\embed(w))$ is defined in \eqref{eq: def weight proba}. The exponential term represents the similarity between the words; words with closely related meanings will hence have a similarity close to one due to the combined use of a common embedding $\embed$ with a smooth kernel.

\begin{remark}
While in theory \eqref{eq: def similarity corpora} includes all the words of both corpora, it is generally-speaking intractable in practice due to the size of the vocabulary of each corpus.
Moreover, since rare words are likely to be unrelated to the text corpus topic and have little influence on the similarity metric (due to their negligible probability weights), we choose to use only the top $500$ most frequent non-stop words of each corpus when computing similarities. 
In our experiments, this value achieved the best trade-off between reducing the computation time and capturing most of the corpus characteristics. 
\end{remark}

\paragraph{Ranking Embeddings} Finally, the ranking of the embeddings is derived from the similarity metric defined in \eqref{eq: def similarity corpora}. For a task associated to a text corpus $\corpus$, we define $\alpha$ by 
\begin{equation}\label{eq: def weight word in task}
\alpha_i(w)= \left\lbrace
\begin{aligned}
&\frac{\# \corpus(w)}{ \sum_{j=1}^N \# \corpus_j(w)  }   \frac{ \sum_{j=1}^N \vert \corpus_j \vert  } {\vert \corpus \vert }\\
& 0 
\end{aligned} \quad  \begin{aligned}
& \text{ if $\exists i \in \left[1,N \right]$} \\ 
& \quad\quad\text{s.t. $w \in \corpus_i$} \\
& \text{ otherwise.}
\end{aligned} \right.
\end{equation}
Note that compared to \eqref{eq: def weight word}, $\alpha(w)$ is the ratio between the frequency of the word $w$ in $\corpus$ and its frequency among the concatenated text corpora of the embeddings, not including $\corpus$. 
Consequently, a word that is present in $\corpus$ but not in any $\corpus_i$ is ignored, as its relative frequency cannot be computed. Then, $P(\embed(w))$ and $s(\corpus_i,\corpus)$ are defined using \eqref{eq: def weight word in task}, \eqref{eq: def weight proba} and \eqref{eq: def similarity corpora}.
Finally, each encoding is ranked according to the similarity between its corpus and the task corpus:
$$ \forall 1\le i\le N,   \quad \quad s(\embed_i,\corpus)\doteq s(\corpus_i,\corpus) $$
The performance of this ranking approach is evaluated in Section \ref{sec:evaluation}.
%In the following, we assume that the embeddings are ordered by decreasing similarity with respect to the task, i.e. 
%$$ s(\embed_1,\corpus) \ge \ldots \ge  s(\embed_N,\corpus). $$

\subsection{Fusing Word Embeddings}\label{subsec: combination}
While in some cases a domain-specific encoding is able to outperform the general-purpose embedding, our experiments show that we can further improve the performance by \emph{merging} multiple domain-specific embeddings (see Section~\ref{sec:evaluation}).
Therefore, the second step of our approach is to combine top-$k$ embeddings from the previous ranking into a new encoding suited for the task at hand. 
We distinguish four different families of embedding combination algorithms: concatenation, averaging, PCA,
%Deep PCA
 and autoencoders, which are discussed in their respective subsections below.

\subsubsection{Concatenation}
The first approach $\embed_{concat}$ simply uses the cartesian product of the different embeddings, i.e. 
$$
\embed_{concat}(\cdot) = \big( \embed_1(\cdot), \ldots, \embed_k(\cdot) \big) \in \mathbb{R}^{kd}. 
$$
By definition, $\embed_{concat}$ retains all semantic information carried by the word in the top-$k$ embeddings. However, this is done at the cost of an increased dimensionality (in this case,  $kd$). Since many of these dimensions may yield redundant -- or non-relevant -- semantic meaning, this increased dimensionality may reduce the efficiency of learning algorithms leveraging $\embed_{concat}$ \citep{bousquet2002stability,vapnik1999overview}. Therefore, the three following algorithms preserve the original dimensionality $d$.  

\subsubsection{Averaging}
Recent approaches have shown that $\embed_{avg}$ constructed from the arithmetic mean between pairs of vectors drawn from distinct $\embed_i$ performs comparably to more complex combination methods under the assumption that the two embeddings are approximately orthogonal~\cite{coates2018frustratingly}.
This method preserves the original dimensionality $d$, however at the cost of a greater loss of information than the following two approaches, which use dimensionality reduction on $\embed_{concat}$ to obtain a final embedding of dimension $d$.

\subsubsection{PCA}\label{sec:pca}
The gold standard for linear methods of dimensionality reduction is the Principal Component Analysis (PCA) \citep{jolliffe2011principal}.
In our setting, we define $\embed_{pca}$ as the result of the PCA algorithm applied on the covariance matrix of the $\embed_{concat}$ embedding of all words in the target corpus $\corpus$. We set the target dimension of $\embed_{pca}$ to $d=300$.

%%%%%%%%%%%%%%%%%%%%%%%%%%%%%%%%%%%%%%%%%%
\iffalse
    \subsubsection{Deep PCA}\label{sec:deeppca}
    We also introduce an extension of the PCA algorithm, called Deep PCA. Deep PCA relies on a layered PCA which improves the performance when combining $k>2$ different embeddings. It works as follows:
    First PCA is used on each pair of embeddings followed by the application of PCA on the cartesian product of these resulting embeddings. The main idea behind Deep PCA is that certain dimensions of the feature vector are more distinctive when compared to a specific, further embedding than when all embeddings are combined and reduced simultaneously.
    % We concatenate and reduce each subpair of desired embeddings and finally combine the outputs from all previous reductions into another PCA layer.

    Formally, given each ordered pair of selected embeddings $(\embed_i, \embed_j)$ with $i>j$, we construct their concatenation $\embed_i(w) \oplus \embed_j(w)$ and apply PCA as in Section~\ref{sec:pca}, resulting in a $d$-dimensional encoding $\embed_{ij}$:
    $$\embed_{ij} = \text{PCA}\big(\embed_i \oplus \embed_j \big).$$
    Then the final $d$-dimensional encoding $\embed_{dpca}$ is obtained by applying PCA to the concatenation of the resulting encodings, with
    %
    $$\embed_{dpca} =  \text{PCA}\big(\bigoplus_{ 1 \le i <j \le k} \embed_{ij} \big)$$
    %
\fi
%%%%%%%%%%%%%%%%%%%%%%%%%%%%%%%%%%%%%%%%%%%

\subsubsection{Autoencoders}\label{sec:othermethods}
The last family of embeddings is obtained by using unsupervised deep learning-based dimensionality reduction algorithms, namely autoencoders \citep{hinton2006fast}, generalized autoencoders \citep{wang2014generalized} and variational autoencoders \citep{kingma2013auto}. By applying these algorithms to the concatenated encoding of the target task $\embed_{concat} \big(\corpus \big)$ with a target dimension of $d=300$, we obtain $\embed_{auto}$, $\embed_{G\,auto}$ and $\embed_{V\,auto}$, respectively. These non-linear methods have been shown to produce interesting results when applied to embedding-related problems (e.g.~\citep{li2015hierarchical}).

\section{Evaluation}
\label{sec:evaluation}
In this section we study in detail the performance of \algo and in particular the pros and cons of the dimensionality reduction methods discussed in Section~\ref{subsec: combination}.
%
%The method introduced in section~\ref{sec:method} has the purpose of optimizing embeddings for their performance on various AI tasks.
%
To evaluate the embeddings produced by \algo, we use the common task of sentiment analysis on user reviews \citep{liu2012survey}, where the learning algorithm is tasked with finding whether the author of a review expresses a positive or negative sentiment towards the object of their review.

\paragraph{Sentiment Analysis on Vector Space Models} As noted by Guggilla et al.~\cite{guggilla2016cnn}, both recurrent neural networks (RNN) and convolutional neural networks (CNN) achieve comparable performance on sentiment analysis 
(see~\cite{ain2017sentiment} for an in-depth review).
\begin{figure*}[tb]
\centering
  \includegraphics[width=.6\textwidth]{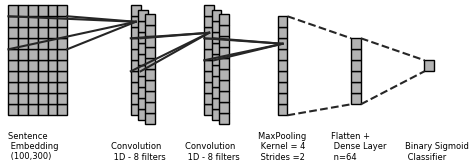}
  \caption{\label{fig:cnn_structure} CNN Architecture used in the evaluation process. Each of the convolutional layers as well as the dense layer is followed by a RELU activation function. } 
\end{figure*}
In this work we selected a CNN architecture for our evaluation as an examplary task. The network takes as input the embeddings of the words of one review and outputs the probability of the review to express a positive sentiment. The architecture of the CNN used in the experiments is summarized in Fig.~\ref{fig:cnn_structure}. For regularization, we used dropout~\citep{srivastava2014dropout} with $p=0.3$ after both the maxpooling layer and the dense layer. The network was trained using Adagrad~\citep{duchi2011adaptive} with batches of 100 sentences over 100 epochs and a learning rate of $0.01$.

\paragraph{Annotated Review Datasets} To evaluate \algo, we used annotated datasets where each review is either labeled as positive/negative or associated with a numeric rating which can be split into positive (upper half of the scale) or negative (lower half of the scale).
We use the following openly available annotated datasets from different domains: movie reviews~\cite{maas2011imdb}, annotated tweets about US airlines\footnote{\url{https://www.kaggle.com/crowdflower/twitter-airline-sentiment}}, Yelp reviews\footnote{\url{https://www.yelp.com/dataset}} including two sets of reviews sampled by selecting reviews within specific categories from the full corpus (\textit{Restaurants} and \textit{Public Services \& Government (PSG)}).
Table~\ref{tab:datasets} gives an overview of these datasets and their cardinality in terms of unique tokens.
\begin{table}[tb]
\centering
\begin{center}
\caption{\label{tab:datasets} Composition of the test datasets consisting of annotated reviews for sentiment analysis and the library of corpora $\corpus$ used to train the embeddings. }
\begin{tabular}{l|r} % TODO: update 
\hline \bf Datasets & \bf \#unique tokens \\ \hline
Movies & 300k \\
Airlines & 17k \\
Full Yelp & 1840k \\
Yelp:Restaurants & 1207k \\
Yelp:PSG & 75k \\
\hline
\\
\bf Embeddings \\
\hline
\textit{Wikipedia} & 2546k \\
Wiki:Actors & 314k  \\
Wiki:Drugs & 246k \\
Wiki:Schools & 348k \\
Wiki:Rail transport & 291k \\
Wiki:Cuisine & 256k \\
Enron & 389k \\
Legal & 83k \\
Twitter & 1920k \\
Subset of Yelp & 38k \\
\hline
\end{tabular}
\end{center}
\end{table}
\paragraph{Domain-Specific Text Corpora} Table~\ref{tab:datasets} also lists the text corpora $\corpus$ which we used to train the domain-specific word2vec embeddings $\embed$.
\textit{Wikipedia} refers to the corpus of all Wikipedia articles.
Corpora labeled \textit{Wiki:} have been obtained by sampling the complete collection of Wikipedia articles, taking into account articles belonging to the given category or its sub-categories only by parsing the Wikipedia category tree. While covering distinct domains, the nature of Wikipedia articles is however such that their language is rather uniform, in order to be understood by a large audience.
The Enron dataset \cite{klimt2004introducing} contains e-mails between employees of the Enron corporation. Given the nature of corporate e-mails, these emails contain mainly colloquial business-related as well as formal communications.
The corpus of legal texts contains court cases from the Federal Court of Australia \cite{lexa} which are written in very specific legal language and terminology. 
We also created an embedding based on a small subset -- $2.5\%$ of the data -- of the full Yelp reviews corpus (abbreviated as \textit{syelp}). In terms of language and terminology, we are expecting this corpus to be similar to other review datasets.
The \textit{twitter} dataset was gathered from the Twitter API by sampling from the open stream over a period of 3 months in early 2016.
The number of unique tokens for these corpora reflects the size of the embeddings' vocabularies, i.e., it only includes words that occurred at least 5 times.

It should be pointed out that the evaluation of \algo is nontrivial, as it requires a large and diverse set of test corpora as well as domain-specific corpora that have to be trained individually and then evaluated against the test corpora.

\subsection{Ranking Embeddings}
First, we evaluate the relevance of the similarity function $s$~\eqref{eq: def similarity corpora} and its induced ranking function introduced in Section~\ref{subsec: ranking}.  Table \ref{tab:topwords} shows the top-3 words according to the $\alpha$ weights~\eqref{eq: def weight word in task} for each embedding.
\begin{table}[tb]
\centering
\begin{center}
\caption{\label{tab:topwords} Top-3 words for each encoding $\corpus_i$, ranked in decreasing order with respect to $\alpha_i$. }
\begin{tabular}{l|l}
\hline
 \bf Text Corpus  &\bf Top-3 words \\ 
 \hline
Actors & protofeminist, swordfights, storywriter  \\ 
Drugs & polydrug, angiography, cyclos \\
Schools & auditoria, schoolmaster, Rabelaisian \\
Rail & southeasternmost, podcar, transponder \\
Cuisine & folktale, ripeness,  anisette \\
Enron & Pinales, rebuttable, customers \\
Legal & hamermill, indictable, mitigator \\
Twitter & queening, punches, theworst \\
Syelp & ambience, chipotle, houseware \\
\hline
\end{tabular}
\end{center}
\end{table}
This highlights that our weight model $\alpha$ is successful in identifying uncommon words that may encode part of the corpus characteristics. However, it also shows that a large number of words may be required to fully capture the aforementioned characteristics.

Table \ref{tab:top2similarity} lists for each dataset the top-2 ranked embeddings, as well as their respective similarity values~\eqref{eq: def similarity corpora} computed with the top $500$ words and $\sigma=0.01$.  We compare our approach to the widely used tf-idf weighted cosine similarity method, \textit{tf-idf-cos} for short (see for example \cite{turney2010frequency}) and against the top-2 embeddings selected manually by an expert, i.e., those that a human expert would expect to be relevant to the domain of the task.
\begin{table}[tb]
\centering
\begin{center}
\caption{\label{tab:top2similarity} Top-2 embeddings automatically selected by the ranking algorithm and by TF-IDF for each test set, with their similarity values, as well as the top-2 expert-picked embeddings for this task. }
\begin{tabular}{l|cc|c|c}
\hline 
\bf Test set  &\bf Automatic  & \bf Similarity $s$ & \bf Manual  & \textbf{tf-idf-cos}\\ 
\bf   &\bf selection & \bf  & \bf selection & \textbf{selection}\\ 
\hline
\multirow{2}{*}{Movies} &  actors &$1.4\times 10^{-4}$ &   actors & syelp \\
&  syelp & $7.5\times 10^{-5}$ & syelp  & twitter\\
\hline
\multirow{2}{*}{Airlines} & twitter & $5.1\times 10^{-6}$ & twitter  & syelp\\
& rail& $1.7\times 10^{-6}$ & rail & twitter\\
\hline
\multirow{2}{*}{Yelp} & syelp &$1.1\times 10^{-3}$ & syelp & syelp \\
&  cuisine & $1.7\times 10^{-5}$& cuisine  & cuisine\\
\hline
\multirow{2}{*}{Restaurants} & syelp&$1.0\times 10^{-3}$ & syelp & cuisine\\
&cuisine &$2.3\times 10^{-4}$ & cuisine & twitter\\
\hline
\multirow{2}{*}{PSG} & syelp &$4.7\times 10^{-4}$ & syelp  & syelp\\
& rail & $2.3\times 10^{-5}$ & legal & cuisine\\
\hline
\end{tabular}
\end{center}
\end{table}
All the selected embeddings match what would be expected from the text corpus domains, except for the \textit{PSG} dataset, for which the \textit{rail} embedding was chosen instead of an intuitively more relevant legal-oriented embedding.

It should be noted that \textit{syelp} is selected in almost every case, which is coherent with the fact that the datasets contain reviews and opinions -- the main subject of the \textit{syelp} corpus. 
Also, the choice of the \textit{twitter} embedding for the \textit{airline} dataset is coherent with the fact that the \textit{airline} corpus is constituted of tweets. 

While the similarity scores in Table \ref{tab:top2similarity} may appear low, they are larger than similarities between unrelated text corpora by several orders of magnitude: For instance, the similarity between \textit{airline} and \textit{actors} is $4.0 \times 10^{-51}$, which highlights the fact that our similarity metric is successful in identifying related text corpora. On the other hand, \textit{tf-idf-cos} always selects a relevant embedding as the first choice, but makes two mistakes for the second choices, potentially related to the highly specialized nature of our text corpus.

\subsection{Domain-Specific Embeddings}\label{sec:domain-specific}
\begin{figure}[tb]
  \centering
    \includegraphics[width=0.375\textwidth]{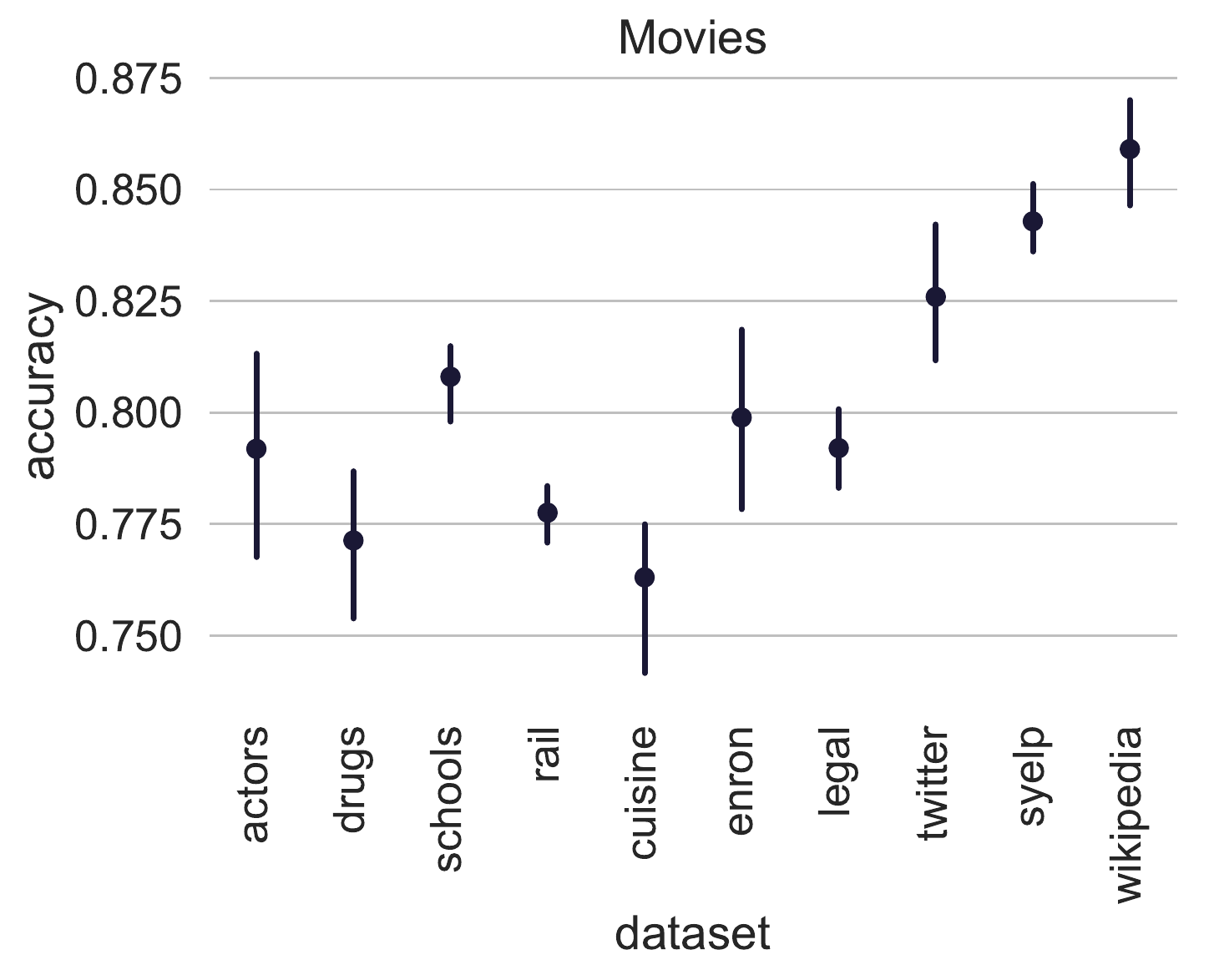}
    \includegraphics[width=0.375\textwidth]{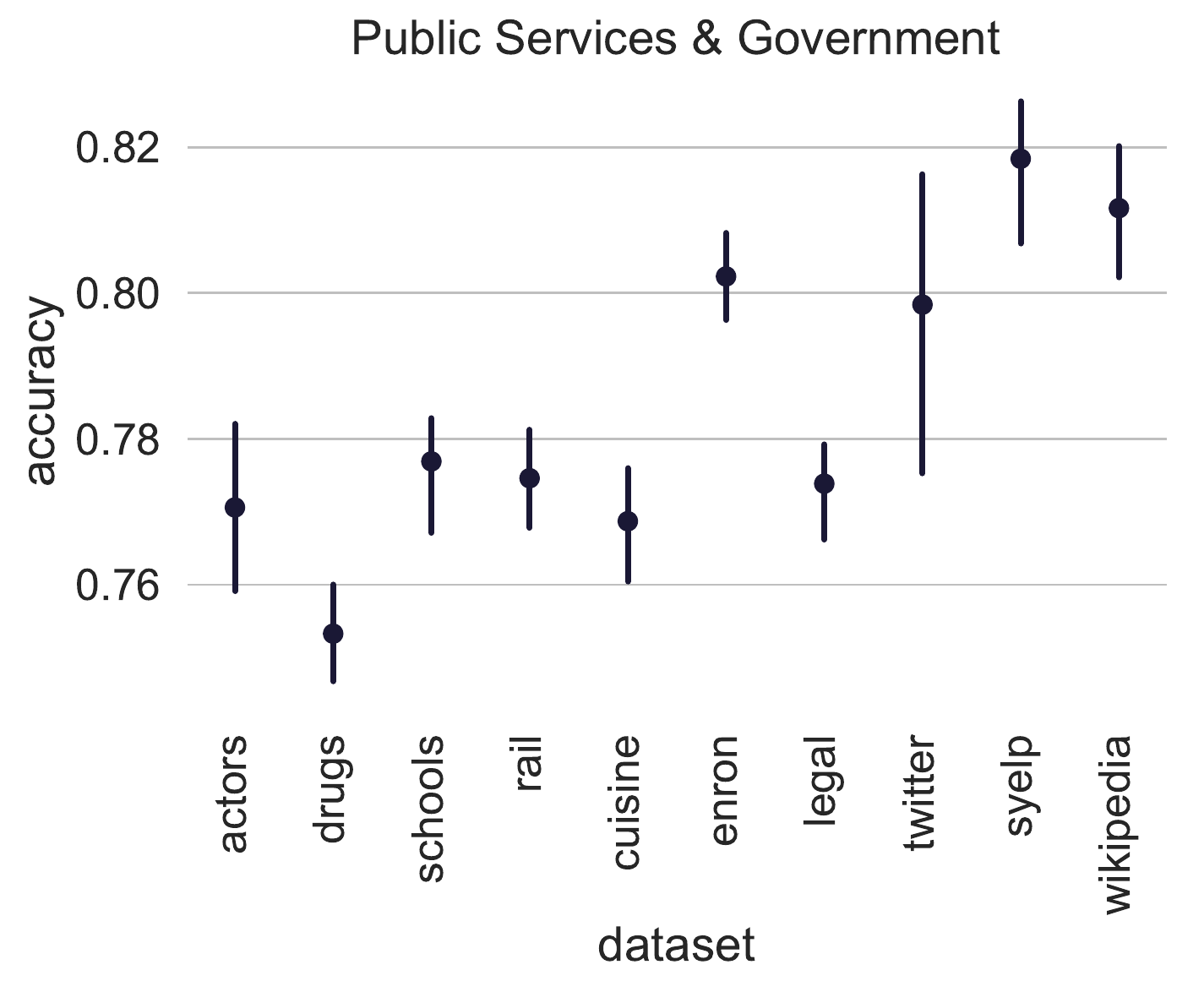}
  \caption{\label{fig:individual_embeddings} Accuracy and standard deviation of all individual domain-specific embeddings compared to the general-purpose \textit{Wikipedia} embedding for two test datasets, \textit{movies} and \textit{PSG}.}
\end{figure}

\begin{table}[tb]
\centering
\begin{center}
\caption{\label{tab:individual_embeddings} Individual domain-specific and general-purpose embeddings' performance for each test dataset.%, compared to general-purpose Wikipedia embeddings. 
}
\begin{tabular}{l|c|c|c|c|c}
\hline \bf Embedding &\bf Movies & \bf Airlines & \bf Yelp & \bf Restaurants & \bf PSG \\ 
\hline
Actors & $0.79$ & $0.86$ & $0.84$ & $0.83$ & $0.77$ \\
% \hline
Drugs & $0.77$ & $0.85$ & $0.79$ & $0.79$ & $0.75$ \\
% \hline
Schools & $0.81$ & $0.87$ & $0.82$ & $0.82$ & $0.78$ \\
% \hline
Rail & $0.78$ & $0.88$ & $0.81$ & $0.81$ & $0.77$ \\
% \hline
Cuisine & $0.76$ & $0.85$ & $0.80$ & $0.82$ & $0.77$ \\
% \hline
Enron & $0.80$ & $0.88$ & $0.82$ & $0.84$ & $0.80$ \\
% \hline
Legal & $0.79$ & $0.86$ & $0.80$ & $0.79$ & $0.77$ \\
% \hline
Twitter & $0.82$ & $\mathbf{0.91}$ & $0.85$ & $0.85$ & $0.79$ \\
% \hline
Syelp & $0.84$ & $0.90$ & $\mathbf{0.86}$ & $\mathbf{0.86}$ & $\mathbf{0.81}$ \\
% \hline
\textit{Wikipedia} & $\mathbf{0.85}$ & $0.90$ & $0.84$ & $0.84$ & $\mathbf{0.81}$ \\
\hline
\end{tabular}
\end{center}
\end{table}
Here we evaluate the individual performance of the topic-specific embeddings. 
We also compare them to general-purpose word2vec encodings trained on the full corpus of English Wikipedia articles. For each of these embeddings, we evaluate the previously defined CNN network and average the results over five runs (the dataset was randomly split into train and test sets (70\%-30\%) for each run). The results are presented in
Table~\ref{tab:individual_embeddings}.
As expected, the general \textit{Wikipedia} embedding is always best or a close second on every task (see also in the plot for the PSG dataset in Fig.~\ref{fig:individual_embeddings}).
On the other hand, the  domain-specific embeddings are worse for most of the tasks in prediction accuracy. 
At the same time, we can observe that the \textit{syelp} embedding performs well on all three Yelp-derived test sets, which is not unexpected given that the embedding was trained on a corpus partially overlapping the test set and is particularly relevant to all opinion-related tasks. %However, the Wikipedia embedding remains a close second, see also in the plot for the PSG dataset in Figure~\ref{fig:individual_embeddings}.
% -- with the exception of the Twitter embedding, which is the closest in domain to the airline review dataset as it consists of tweets.
These results highlight the need of both the ranking function, to select relevant domain specific embeddings, as well as the combination of these embeddings.

\subsection{Embedding Fusion}\label{sec:embedding-fusion}
\begin{table*}[tb]
\centering
\begin{center}
\caption{\label{tab:dim_red_methods} Comparison of dimensionality reduction methods for the top-2 automatically selected text corpora for all given tasks. The first row reports average accuracy, the second row the standard deviation over 5 runs.% The last column shows the performance using the Wikipedia embeddings for comparison.
}
\begin{tabular}{l|c|c|c|c|c|c|c|c}
\hline \bf Test set  &\bf Corpora & $\embed_{concat}$ & $\embed_{PCA}$ & $\embed_{auto}$ & $\embed_{G\,auto}$ & $\embed_{V\,auto}$ & \textit{Wikipedia} & $\embed_{avg}$\\ 
\hline
Movies & actors, syelp & $0.84$ & $\mathbf{0.87}$ & $0.86$ & $0.82$ & $0.83$ & $0.85$ & $0.84$  \\
 & & $0.013$ & $0.0052$ & $0.017$ & $0.015$ & $0.014$ & $0.013$ & $0.011$ \\
\hline
Airlines & twitter, rail & $\mathbf{0.91}$ & $\mathbf{0.91}$ & $0.90$ & $0.86$ & $0.90$ & $0.90$ & $0.90$ \\
 & & $0.013$ & $0.0071$ & $0.0074$ & $0.016$ & $0.012$ & $0.0081$ & $0.020$  \\
\hline
Yelp & syelp, cuisine & $0.86$ & $0.86$ &  $\mathbf{0.87}$ & $0.86$ & $0.86$ & $0.84$ & $0.86$ \\
 & & $0.0037$ & $0.0012$ & $0.0023$ & $0.0044$ & $0.0021$ & $0.013$ & $0.0040$ \\
\hline
Restaurants & syelp, cuisine & $0.85$ & $\mathbf{0.86}$ &  $\mathbf{0.86}$ & $0.85$ & $0.85$ & $0.84$ & $\mathbf{0.86}$ \\
 & & $0.012$ & $0.0014$ & $0.0026$ & $0.0025$ & $0.0046$ & $0.0096$ & $0.0028$  \\
\hline
PSG & syelp, rail & $0.81$ & $\mathbf{0.83}$ &  $0.82$ & $0.80$ &  $0.81$ & $0.81$ & $0.82$ \\
 & & $0.012$ & $0.0098$ & $0.0079$ & $0.0082$ & $0.0058$ & $0.010$ & $0.0033$ \\
\hline
\end{tabular}
\end{center}
\end{table*}
We now evaluate the performance of the different dimensionality reduction methods described in Section~\ref{subsec: combination} for the different datasets, using the two top-ranked domain-specific embeddings (cf. Table~\ref{tab:top2similarity}) in all cases. 
For each of the aforementioned embeddings, we perform five runs of the CNN network (with a random split (70\%-30\%) for train and test sets).
% For comparison purposes, we also run the CNN with the FastText embedding, as well as with both domain-specific embeddings individually.
The results are reported in Table~\ref{tab:dim_red_methods} and compared against $\embed_{avg}$~\cite{coates2018frustratingly} as well as the general-purpose \textit{Wikipedia} word2vec model used individually.
\begin{figure}[tb]
  \centering
  \includegraphics[width=0.375\textwidth]{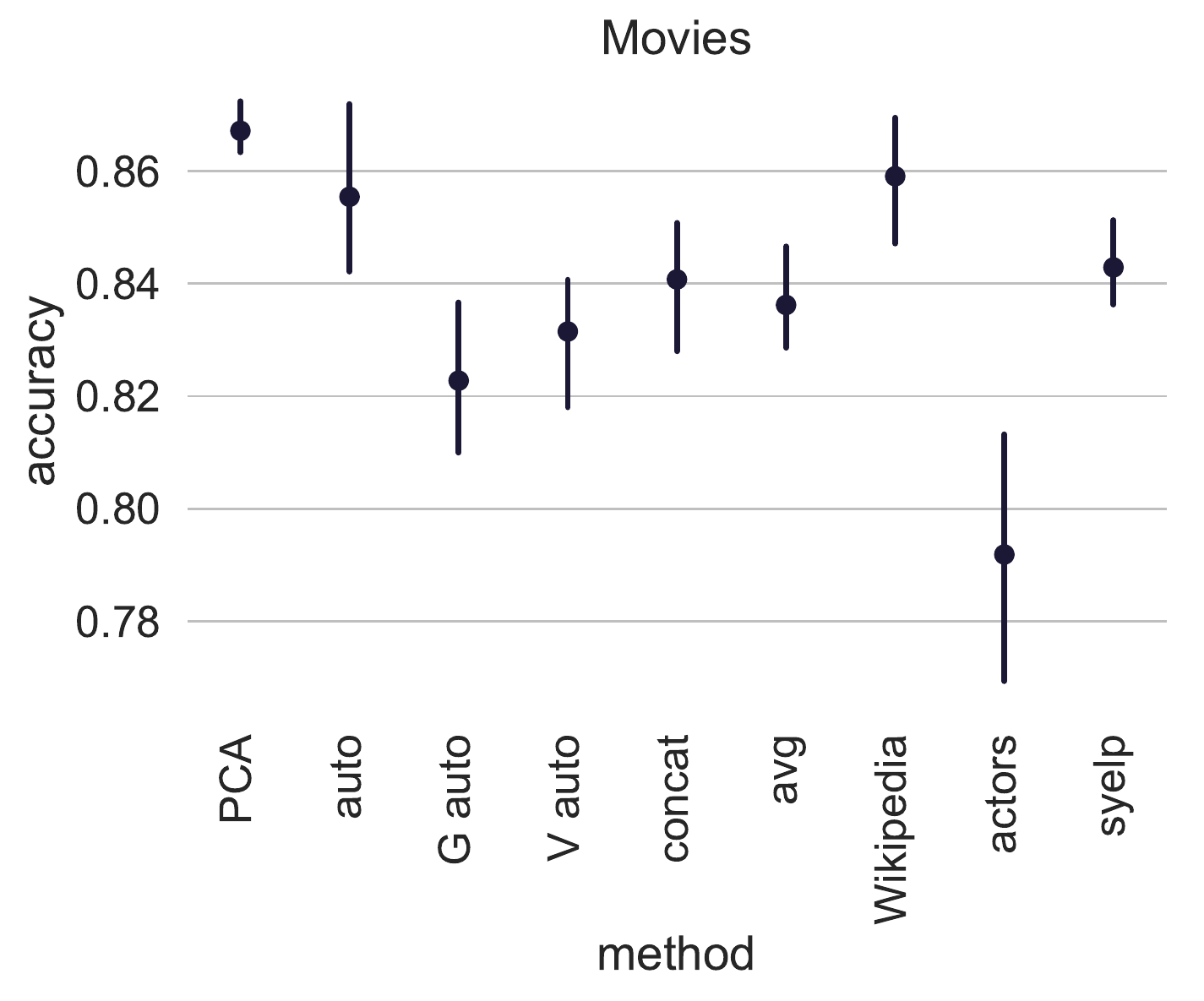}
  \includegraphics[width=0.375\textwidth]{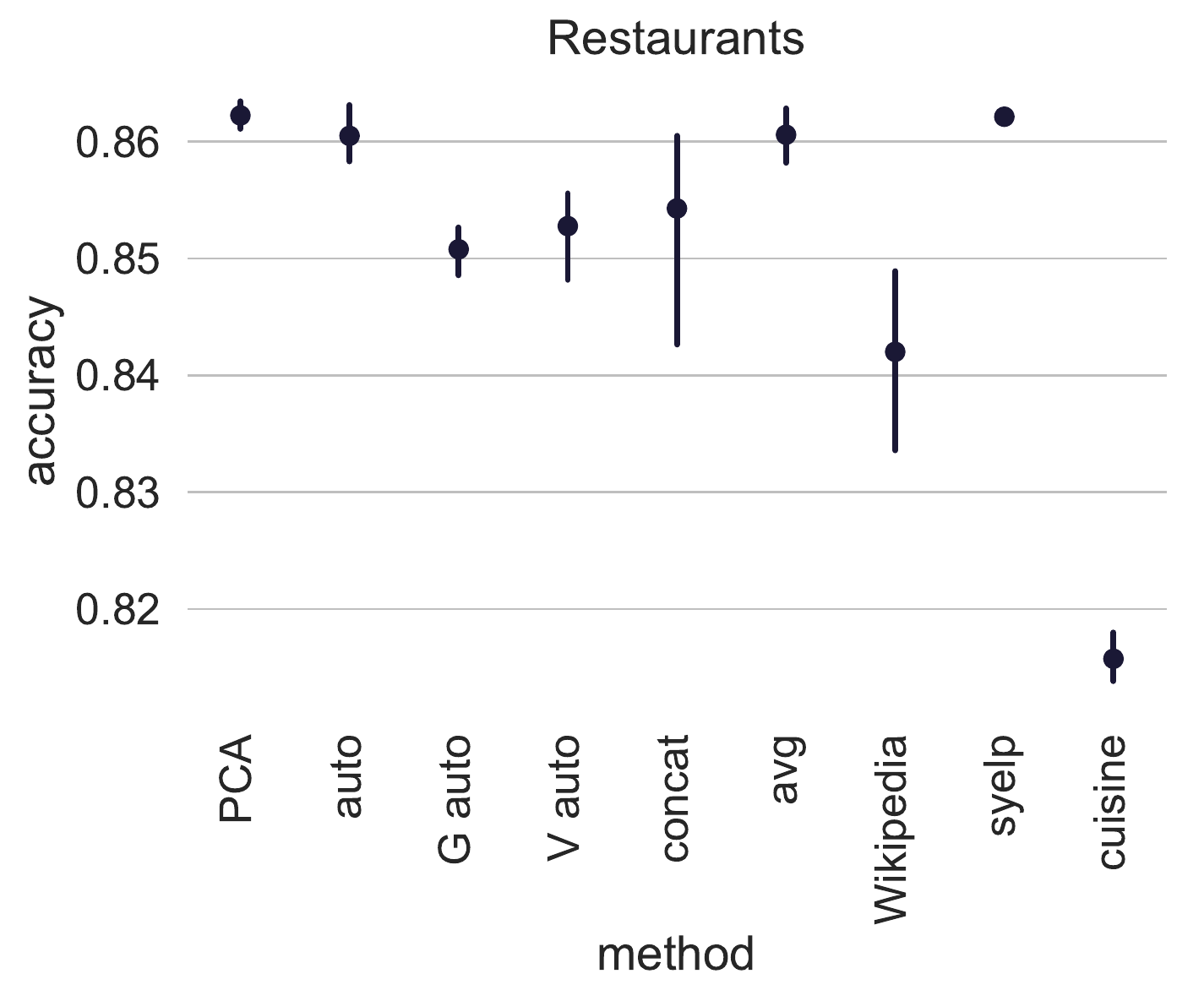}
  \caption{\label{fig:dim_red_methods} Comparison of the accuracy and standard deviation of individual embeddings and fused embeddings obtained by the different dimensionality reduction methods on the movie review  and restaurant review datasets.}
\end{figure}
The results show that among the different embedding reduction methods, $\embed_{PCA}$ 
%(which is equal to $\embed_{DPCA}$ in this case, as there are only two domain-specific encodings to combine)
outperforms the other methods -- although $\embed_{auto}$ is a close second, slightly outperforming $\embed_{PCA}$ on the Yelp review test set. $\embed_{PCA}$ also performs slightly better than \textit{Wikipedia}, even though $\embed_{PCA}$ was trained on text corpora that are smaller by several orders of magnitude.
Overall, $\embed_{PCA}$ never deteriorates the performance of the individual embeddings and  outperforms both individual embeddings by a significant margin in two cases (\textit{Movies} and \textit{PSG}).

\section{Discussion}
\label{sec:discussion}
\paragraph{Domain-Specific Embeddings} As can be seen from Section~\ref{sec:domain-specific}, domain-specific encodings tend to be less effective than general-purpose embeddings when used individually. This is in line with the fact that encodings based on algorithms such as word2vec are typically trained using very large text corpora and therefore contain extensive vocabulary as well as information-rich encodings for many words due to their repeated occurrence in text. 
As can be seen in both Table~\ref{tab:individual_embeddings} and Fig.~\ref{fig:individual_embeddings}, the individual embedding built from the subsampled set of Yelp reviews performs better than the general-purpose \textit{Wikipedia} embedding for the three Yelp-derived test sets. This is easy to explain by the fact that there is a very high similarity, potentially even an overlap, between training and test data. We nevertheless chose to include the \emph{syelp} embedding in the tests as it performed well also on the non-Yelp test sets (cf. Table~\ref{tab:individual_embeddings}).
While we consistently used the well-established word2vec algorithm to train our set of domain-specific embeddings, we believe our method translates equally to embeddings obtained through more recently proposed methods~\cite{peters2018deep,bojanowski2016enriching}. % maybe move this sentence to another section such as conclusion?

\paragraph{Combining Domain-Specific Embeddings} The combination of relevant domain-specific embeddings was shown to consistently outperform the general-purpose embedding (see Section~\ref{sec:embedding-fusion}) --  despite using significantly smaller text corpora. This might be explained by the fact that a very large text corpus may induce a loss of information for some words that have different meanings in specific contexts, a loss that is also present with $\embed_{avg}$. Indeed, embeddings might ``average'' those meanings during their training %, which would lead to a loss of information 
-- a problem that is not encountered by domain-specific encodings due to their narrower usage of specific terms. These semantics are preserved in an embedding that is combined at a later stage as opposed to trained on a concatenation of the original corpora, where this averaging of meanings still occurs.

\paragraph{Ranking Embeddings} Table~\ref{tab:top2similarity} shows that our ranking function selects the same embeddings as in the manual selection by human experts in all cases except one (PSG). 
It is interesting to note that the selected encodings include \emph{actors} and \emph{cuisine} -- corpora related to the topic of the task -- but also \emph{twitter} -- a general corpus related to the medium and language of the task -- and \emph{syelp} -- a corpus related to the content of the task, i.e., opinions.
This highlights the generality and the multi-purpose usefulness of our similarity score and offers a promising first step in devising a more refined algorithm that combines representative encodings derived from three separate libraries of embeddings: topic, medium, and content.

%  a choice that reflected the subject of in the manual selection -- the ranking also tends to predict \emph{syelp} and \emph{twitter} as significantly similar to the target corpus. These embeddings are related to the target topic (syelp is a corpus of opinions, and twitter contains a higher than average proportion of opinion words), but are not as good as the handpicked embeddings, as illustrated by the CNN performance. Two reasons might explain this result. First, in our model, text copora are only described by a small number of frequent words (100 in our case). This small sample may contain redundant words, and may lack significant yet less frequent words, and therefore the similarity function might greatly benefit from the use of a custom-made list of words. Second, this approach only takes words and their frequency into account. It is the authors' opinion that a more reliable similarity measure might be deduced by extracting entities and relations from each corpus, and comparing the resulting knowledge graphs.

\paragraph{Dimension Reduction} According to Table~\ref{tab:dim_red_methods} and Fig.~\ref{fig:dim_red_methods}, linear dimensionality reduction methods (i.e., PCA-based approaches) perform better than nonlinear methods (in this case, autoencoders and in particular their extensions) reliably, i.e., with a comparably low standard deviation. This result might be explained by the fact that Principal Component Analysis extracts the directions that account for the largest amount of variation. Since differences between words (i.e., variations) have been shown to be linked to relations~\citep{levy2014linguistic}, PCA may extract directions that are linked to important relations for this particular topic -- since the different corpora have been created to be domain-specific. %Further work is required to fully understand the nature of PCA performances.

%\paragraph{Deep PCA} Our proposed Deep PCA method performs similarly or better than a ``flat PCA'', as can be observed in Table~\ref{tab:deeppca}. It is our understanding that this improvement can be attributed, similar to the results using PCA, to the selection of directions with the largest amount of variation between each pair of involved embeddings. A Deep PCA approach enhances the variation on individual pairs of embeddings before combining all embeddings into a joint representation. On the other hand, the fact that Deep PCA does not always lead to an improvement can be explained by the fact that a one-step PCA already performs well at selecting distinctive dimensions when applied over all simultaneously. 
%However, while we did not formally evaluate the runtime performance of \algo, we observed that our layered PCA approach is not significantly slower than a direct PCA approach. %, supporting the efficiency of the method.

\section{Related Work}
\label{sec:relwork}
\paragraph{Multi-Corpus Word Embeddings}

One common NLP task using two or more embeddings is machine translation, where there is one embedding per language. Methods are often put in place to align the vector spaces in order to find matching words without knowing the full ground-truth \cite{lample2018phrase,yang2013word}. 
When not performing alignment, machine translation requires parallel or aligned text corpora \cite{gouws2015bilbowa,vulic2015bilingual}. Unlike these methods, our approach does not require the vector spaces to be aligned; however, since we are dealing with monolingual cases, our corpora are implicitly parallel.
Another case for joining embeddings is when dealing with joint embeddings of different types of data, such as knowledge graphs with text \cite{zhong2015aligning}, or images with text \cite{norouzi2013zero}. While this is in essence an alignment between different structures, our method operates solely on textual contents and requires corpora with a textual overlap.
State-of-the-art approaches to multi-domain embeddings~\cite{yang2017simple} require a joint training of the embeddings into a common space. Our method does not require retraining when choosing to combine embeddings of certain domains.

\paragraph{Combining Word Embeddings} Previous works have proposed different approaches to combine word embeddings, the most fundamental being the concatenation of embeddings trained with different algorithms~\cite{ghannay2016word}. However, this work does not consider domain-specific embeddings. 
Similarly, the method of averaging embeddings~\cite{coates2018frustratingly} creates a ``meta-embedding'' from the arithmetic mean of the vectors that is comparable in accuracy to concatenation of vectors while offering the performance benefits of lower-dimensional embeddings. We have shown to outperform this method in Table~\ref{tab:dim_red_methods}.
Yin and Sch\"utze~\cite{yin2015multichannel} use a multi-channel approach to word embedding, borrowing from image processing techniques and using a different encoding for each channel, while Zhang et al.~\cite{zhang2016mgnc} compute the first layers of the CNN with different word embeddings in parallel and concatenate them at the very last layer. 
However, and to the best of our knowledge, we are the first to propose the idea of dynamically selecting and combining several domain-specific word embeddings.

\paragraph{Sentiment Analysis} Sentiment analysis, and in particular predicting whether a given text conveys a positive or negative message, has been the subject of many publications in the recent past \citep{liu2012survey}. Deep learning networks have been a tool of choice for this task, particularly when combined with word embeddings \citep{kim2014convolutional}. While both recurrent neural networks and convolutional neural networks have been successfully used for this task, both families have been shown to achieve similar performance \citep{yin2017comparative,ain2017sentiment}. 
Tang et al.~\cite{Tang2016SentimentEW} introduced the use of specific sentiment embeddings, i.e. embeddings of words with their sentiment, for the specific task of sentiment analysis. This is not suitable for our approach, as we intend to generalize to various AI tasks.

\paragraph{Dimensionality Reduction}
Dimensionality reduction is a key component of many statistical learning approaches~\citep{sorzano2014survey}. 
The combination of concatenation followed by dimensionality reduction is not new, as it is at the heart of many tensor-based learning methods~\citep{rabusseau2017multitask}. In these cases, the dimension reduction is achieved through the choice of an appropriate regularization, such as constraints on the CP rank of the tensor~\citep{goldfarb2014robust} -- an approach that can be related to the construction of the PCA embedding in our case. Similarly, the use of autoencoders as the first layers of deep neural networks is a common approach when the data are too complex (such as living in a very high-dimensional space)~\citep{hinton2006fast}. 
%There have been previous works which layer PCA with other methods for the task of face recognition~\cite{liong2013face,chan2015pcanet,tian2015stacked}, where PCA is applied to visual features and is interspersed with other feature transformation techniques, unlike the technique proposed here. % don't need deep PCA references
To the authors' knowledge,
% the Deep PCA approach introduced in this paper is novel in its application to embeddings and its usage on its own; moreover,
 none of the methods proposed in this paper have been previously used to combine embeddings.

\section{Conclusions}
\label{sec:conclusions}
In this paper, we introduced the idea of combining individual embeddings to capture domain-specific semantics. In that sense, we presented \algo, a two-step process consisting in ranking and then combining domain-specific embeddings. 
Our ranking method captures the similarity between the corpus used by the downstream application and the various domain-specific embeddings that are available. \algo was shown to select embeddings that are highly relevant and of which the combined performance is higher than a general-purpose embedding.
We showed that the best performing combination method was a PCA approach, $\embed_{PCA}$, that fuses different embeddings into a single efficient and effective embedding, which outperforms each of the  embeddings taken individually as well as its nonlinear counterparts. Compared to the general-purpose \textit{Wikipedia} embedding, $\embed_{PCA}$ yields a consistent and significant performance improvement (2\% improvement on average on already highly-accurate scores) despite being trained on data that is several orders of magnitude smaller. 

In future work, we plan to improve our ranking method by incorporating additional information, e.g., by linking both the application and embedding corpora to a knowledge graph in order to capture their semantic overlap more precisely, and to develop a new approach to effectively combine three or more embeddings by using their characteristics.
For this, we intend to investigate the possibility of combining embeddings that reflect specific properties of a target corpus, nameley, topic, medium, and content from different libraries of embeddings.
 % and improve the ranking function.
%such that it selects the most suitable among the candidate embeddings not only based on corpus similarity, but also based on their performance for the task at hand, which at times may not be correlated with corpus similarity.

\section*{Acknowledgment}

This work was funded by the \emph{Hasler Foundation} in the context of the \emph{City-Stories} project.

\bibliographystyle{IEEEtran}
\bibliography{IEEEabrv,paper}

\begin{thebibliography}{10}
\providecommand{\url}[1]{#1}
\csname url@samestyle\endcsname
\providecommand{\newblock}{\relax}
\providecommand{\bibinfo}[2]{#2}
\providecommand{\BIBentrySTDinterwordspacing}{\spaceskip=0pt\relax}
\providecommand{\BIBentryALTinterwordstretchfactor}{4}
\providecommand{\BIBentryALTinterwordspacing}{\spaceskip=\fontdimen2\font plus
\BIBentryALTinterwordstretchfactor\fontdimen3\font minus
  \fontdimen4\font\relax}
\providecommand{\BIBforeignlanguage}[2]{{%
\expandafter\ifx\csname l@#1\endcsname\relax
\typeout{** WARNING: IEEEtran.bst: No hyphenation pattern has been}%
\typeout{** loaded for the language `#1'. Using the pattern for}%
\typeout{** the default language instead.}%
\else
\language=\csname l@#1\endcsname
\fi
#2}}
\providecommand{\BIBdecl}{\relax}
\BIBdecl

\bibitem{Mikolov:2013}
T.~Mikolov, I.~Sutskever, K.~Chen, G.~Corrado, and J.~Dean, ``Distributed
  representations of words and phrases and their compositionality,'' in
  \emph{Proceedings of the 26th International Conference on Neural Information
  Processing Systems}, 2013, pp. 3111--3119.

\bibitem{hamilton2016inducing}
W.~L. Hamilton, K.~Clark, J.~Leskovec, and D.~Jurafsky, ``Inducing
  domain-specific sentiment lexicons from unlabeled corpora,'' in
  \emph{Proceedings of the Conference on Empirical Methods in Natural Language
  Processing}, 2016, p. 595.

\bibitem{Xu2018LifelongDW}
H.~Xu, B.~Liu, L.~Shu, and P.~Yu, ``Lifelong domain word embedding via
  meta-learning,'' in \emph{International Joint Conference on Artificial
  Intelligence}, 2018.

\bibitem{mikolov2013efficient}
T.~Mikolov, K.~Chen, G.~S. Corrado, and J.~Dean, ``Efficient estimation of word
  representations in vector space,'' \emph{CoRR}, vol. abs/1301.3781, 2013.

\bibitem{bojanowski2016enriching}
P.~Bojanowski, E.~Grave, A.~Joulin, and T.~Mikolov, ``Enriching word vectors
  with subword information,'' \emph{Transactions of the Association for
  Computational Linguistics}, vol.~5, pp. 135--146, 2017.

\bibitem{bousquet2002stability}
O.~Bousquet and A.~Elisseeff, ``Stability and generalization,'' \emph{Journal
  of machine learning research}, vol.~2, pp. 499--526, 2002.

\bibitem{vapnik1999overview}
V.~N. Vapnik, ``An overview of statistical learning theory,'' \emph{IEEE
  transactions on neural networks}, vol.~10, no.~5, pp. 988--999, 1999.

\bibitem{jolliffe2011principal}
I.~Jolliffe, ``Principal component analysis,'' in \emph{International
  encyclopedia of statistical science}.\hskip 1em plus 0.5em minus 0.4em\relax
  Springer, 2011, pp. 1094--1096.

\bibitem{hinton2006fast}
G.~E. Hinton, S.~Osindero, and Y.-W. Teh, ``A fast learning algorithm for deep
  belief nets,'' \emph{Neural computation}, vol.~18, no.~7, pp. 1527--1554,
  2006.

\bibitem{wang2014generalized}
W.~Wang, Y.~Huang, Y.~Wang, and L.~Wang, ``Generalized autoencoder: A neural
  network framework for dimensionality reduction,'' in \emph{Proceedings of the
  IEEE conference on computer vision and pattern recognition workshops}, 2014,
  pp. 490--497.

\bibitem{kingma2013auto}
D.~P. Kingma and M.~Welling, ``Auto-encoding variational bayes,'' in
  \emph{Proceedings of the 2nd International Conference on Learning
  Representations (ICLR)}, 2014.

\bibitem{li2015hierarchical}
J.~Li, T.~Luong, and D.~Jurafsky, ``A hierarchical neural autoencoder for
  paragraphs and documents,'' in \emph{Proceedings of the 53rd Annual Meeting
  of the Association for Computational Linguistics and the 7th International
  Joint Conference on Natural Language Processing (Volume 1: Long Papers)},
  2015, pp. 1106--1115.

\bibitem{liu2012survey}
B.~Liu and L.~Zhang, ``A survey of opinion mining and sentiment analysis,'' in
  \emph{Mining text data}.\hskip 1em plus 0.5em minus 0.4em\relax Springer,
  2012, pp. 415--463.

\bibitem{guggilla2016cnn}
C.~Guggilla, T.~Miller, and I.~Gurevych, ``Cnn-and lstm-based claim
  classification in online user comments,'' in \emph{Proceedings of COLING
  2016, the 26th International Conference on Computational Linguistics:
  Technical Papers}, 2016, pp. 2740--2751.

\bibitem{ain2017sentiment}
Q.~T. Ain, M.~Ali, A.~Riaz, A.~Noureen, M.~Kamran, B.~Hayat, and A.~Rehman,
  ``Sentiment analysis using deep learning techniques: a review,'' \emph{Int J
  Adv Comput Sci Appl}, vol.~8, no.~6, p. 424, 2017.

\bibitem{srivastava2014dropout}
N.~Srivastava, G.~Hinton, A.~Krizhevsky, I.~Sutskever, and R.~Salakhutdinov,
  ``Dropout: a simple way to prevent neural networks from overfitting,''
  \emph{Journal of Machine Learning Research}, vol.~15, no.~1, pp. 1929--1958,
  2014.

\bibitem{duchi2011adaptive}
J.~Duchi, E.~Hazan, and Y.~Singer, ``Adaptive subgradient methods for online
  learning and stochastic optimization,'' \emph{Journal of Machine Learning
  Research}, vol.~12, no. Jul, pp. 2121--2159, 2011.

\bibitem{maas2011imdb}
A.~L. Maas, R.~E. Daly, P.~T. Pham, D.~Huang, A.~Y. Ng, and C.~Potts,
  ``Learning word vectors for sentiment analysis,'' in \emph{Proceedings of the
  49th Annual Meeting of the Association for Computational Linguistics: Human
  Language Technologies}, 2011, pp. 142--150.

\bibitem{klimt2004introducing}
B.~Klimt and Y.~Yang, ``Introducing the {E}nron corpus.'' in \emph{CEAS}, 2004.

\bibitem{lexa}
F.~Galgani and A.~Hoffmann, ``Lexa: Towards automatic legal citation
  classification,'' in \emph{AI 2010: Advances in Artificial Intelligence},
  ser. Lecture Notes in Computer Science, J.~Li, Ed., vol. 6464.\hskip 1em plus
  0.5em minus 0.4em\relax Springer Berlin Heidelberg, 2010, pp. 445 --454.

\bibitem{turney2010frequency}
P.~D. Turney and P.~Pantel, ``From frequency to meaning: Vector space models of
  semantics,'' \emph{Journal of artificial intelligence research}, vol.~37, pp.
  141--188, 2010.

\bibitem{coates2018frustratingly}
J.~Coates and D.~Bollegala, ``Frustratingly easy meta-embedding {--} computing
  meta-embeddings by averaging source word embeddings,'' in \emph{Proceedings
  of the 2018 Conference of the North American Chapter of the Association for
  Computational Linguistics: Human Language Technologies, Volume 2 (Short
  Papers)}, 2018.

\bibitem{peters2018deep}
M.~Peters, M.~Neumann, M.~Iyyer, M.~Gardner, C.~Clark, K.~Lee, and
  L.~Zettlemoyer, ``Deep contextualized word representations,'' in
  \emph{Proceedings of the 2018 Conference of the North American Chapter of the
  Association for Computational Linguistics: Human Language Technologies,
  Volume 1 (Long Papers)}, 2018, pp. 2227--2237.

\bibitem{levy2014linguistic}
O.~Levy and Y.~Goldberg, ``Linguistic regularities in sparse and explicit word
  representations,'' in \emph{Proceedings of the eighteenth conference on
  computational natural language learning}, 2014, pp. 171--180.

\bibitem{lample2018phrase}
G.~Lample, M.~Ott, A.~Conneau, L.~Denoyer \emph{et~al.}, ``Phrase-based \&
  neural unsupervised machine translation,'' in \emph{Proceedings of the 2018
  Conference on Empirical Methods in Natural Language Processing}, 2018, pp.
  5039--5049.

\bibitem{yang2013word}
N.~Yang, S.~Liu, M.~Li, M.~Zhou, and N.~Yu, ``Word alignment modeling with
  context dependent deep neural network,'' in \emph{Proceedings of the 51st
  Annual Meeting of the Association for Computational Linguistics}, 2013, pp.
  166--175.

\bibitem{gouws2015bilbowa}
S.~Gouws, Y.~Bengio, and G.~Corrado, ``Bilbowa: Fast bilingual distributed
  representations without word alignments,'' in \emph{International Conference
  on Machine Learning}, 2015, pp. 748--756.

\bibitem{vulic2015bilingual}
I.~Vulic and M.-F. Moens, ``Bilingual word embeddings from non-parallel
  document-aligned data applied to bilingual lexicon induction,'' in
  \emph{Proceedings of the 53rd Annual Meeting of the Association for
  Computational Linguistics (ACL 2015)}, vol.~2.\hskip 1em plus 0.5em minus
  0.4em\relax ACL; East Stroudsburg, PA, 2015, pp. 719--725.

\bibitem{zhong2015aligning}
H.~Zhong, J.~Zhang, Z.~Wang, H.~Wan, and Z.~Chen, ``Aligning knowledge and text
  embeddings by entity descriptions,'' in \emph{Proceedings of the 2015
  Conference on Empirical Methods in Natural Language Processing}, 2015, pp.
  267--272.

\bibitem{norouzi2013zero}
M.~Norouzi, T.~Mikolov, S.~Bengio, Y.~Singer, J.~Shlens, A.~Frome, G.~S.
  Corrado, and J.~Dean, ``Zero-shot learning by convex combination of semantic
  embeddings,'' in \emph{Proceedings of the 2nd International Conference on
  Learning Representations (ICLR)}, 2014.

\bibitem{yang2017simple}
W.~Yang, W.~Lu, and V.~Zheng, ``A simple regularization-based algorithm for
  learning cross-domain word embeddings,'' in \emph{Proceedings of the 2017
  Conference on Empirical Methods in Natural Language Processing}, 2017, pp.
  2898--2904.

\bibitem{ghannay2016word}
S.~Ghannay, B.~Favre, Y.~Esteve, and N.~Camelin, ``Word embedding evaluation
  and combination,'' in \emph{Proceedings of the Tenth International Conference
  on Language Resources and Evaluation (LREC 2016)}, 2016, pp. 300--305.

\bibitem{yin2015multichannel}
W.~Yin and H.~Sch{\"u}tze, ``Multichannel variable-size convolution for
  sentence classification,'' in \emph{Proceedings of the Nineteenth Conference
  on Computational Natural Language Learning}, 2015, pp. 204--214.

\bibitem{zhang2016mgnc}
Y.~Zhang, S.~Roller, and B.~C. Wallace, ``{MGNC-CNN}: A simple approach to
  exploiting multiple word embeddings for sentence classification,'' in
  \emph{Proceedings of the 2016 Conference of the North American Chapter of the
  Association for Computational Linguistics: Human Language Technologies},
  2016, pp. 1522--1527.

\bibitem{kim2014convolutional}
Y.~Kim, ``Convolutional neural networks for sentence classification,'' in
  \emph{Proceedings of the 2014 Conference on Empirical Methods in Natural
  Language Processing (EMNLP)}, 2014, pp. 1746--1751.

\bibitem{yin2017comparative}
W.~Yin, K.~Kann, M.~Yu, and H.~Sch{\"u}tze, ``Comparative study of cnn and rnn
  for natural language processing,'' \emph{CoRR}, vol. abs/1702.01923, 2017.

\bibitem{Tang2016SentimentEW}
D.~Tang, F.~Wei, B.~Qin, N.~Yang, T.~Liu, and M.~Zhou, ``Sentiment embeddings
  with applications to sentiment analysis,'' \emph{IEEE Transactions on
  Knowledge and Data Engineering}, vol.~28, pp. 496--509, 2016.

\bibitem{sorzano2014survey}
C.~O.~S. Sorzano, J.~Vargas, and A.~D. Pascual-Montano, ``A survey of
  dimensionality reduction techniques,'' \emph{CoRR}, vol. abs/1403.2877, 2014.

\bibitem{rabusseau2017multitask}
G.~Rabusseau, B.~Balle, and J.~Pineau, ``Multitask spectral learning of
  weighted automata,'' in \emph{Advances in Neural Information Processing
  Systems}, 2017, pp. 2588--2597.

\bibitem{goldfarb2014robust}
D.~Goldfarb and Z.~Qin, ``Robust low-rank tensor recovery: Models and
  algorithms,'' \emph{SIAM Journal on Matrix Analysis and Applications},
  vol.~35, no.~1, pp. 225--253, 2014.

\end{thebibliography}

\end{document}